\documentclass{esann}
\usepackage[dvips]{graphicx}
\usepackage[latin1]{inputenc}
\usepackage{amssymb}
\usepackage{subcaption}
\usepackage{booktabs}
\usepackage{tikz}

\newtheorem{definition}{Definition}

\usetikzlibrary{decorations.pathreplacing,calc,shapes,positioning}

\usetikzlibrary{shapes,shadows,arrows,positioning,angles,quotes}
\tikzset{My Arrow Style/.style={single arrow, fill=black!15, anchor=base, align=center,text width=2.3cm}}
\tikzstyle{arrow} = [thick,->,>=stealth]
\usetikzlibrary{calc,trees,positioning,arrows,fit,shapes,calc}

\tikzstyle{startstop} = [rectangle, rounded corners, minimum width=1.5cm, minimum height=0.5cm,text centered, draw=black, fill=red!30]
\tikzstyle{io} = [trapezium, trapezium left angle=70, trapezium right angle=110, minimum width=1cm, minimum height=0.5cm, text centered, draw=black, fill=blue!30]

\tikzstyle{process} = [rectangle, minimum width=3cm, minimum height=0.5cm, text centered, draw=black, fill=orange!30]
\tikzstyle{decision} = [diamond, minimum width=0.5cm, minimum height=0.1cm, text centered, draw=black, fill=green!30]
\tikzstyle{process2} = [rectangle, minimum width=1cm, minimum height=0.5cm, text centered, draw=black, fill=orange!30]
\tikzstyle{arrow} = [thick,->,>=stealth]

\begin{document}
\title{Safety-Oriented Pruning and Interpretation of Reinforcement Learning Policies\footnote{Funded by the EU under grant agreement number 101091783 (MARS Project) and as part of the Horizon Europe HORIZON-CL4-2022-TWIN-TRANSITION-01-03.}}

\author{Dennis Gross and Helge Spieker
%
\vspace{.3cm}\\
%
Simula Research Laboratory \\
Oslo, Norway
}

\maketitle

\begin{abstract}
Pruning neural networks (NNs) can streamline them but risks removing vital parameters from safe reinforcement learning (RL) policies.
We introduce an interpretable RL method called \emph{VERINTER}, which combines NN pruning with model checking to ensure interpretable RL safety.
VERINTER \emph{exactly} quantifies the effects of pruning and the impact of neural connections on complex safety properties by analyzing changes in safety measurements.
This method maintains safety in pruned RL policies and enhances understanding of their safety dynamics, which has proven effective in multiple RL settings.
\end{abstract}

\section{Introduction}
\emph{Reinforcement learning (RL)} has transformed technology~\cite{mnih2013playing}. An RL agent learns a \emph{policy} to achieve a set objective by acting and receiving rewards and observations from an environment. 
A \emph{neural network (NN)} typically represents the policy, mapping environment state observations to action choices.
Each observation comprises features characterizing the current environment state~\cite{DBLP:conf/setta/Gross22}.

Unfortunately, learned policies are not guaranteed to avoid \emph{unsafe behavior}~\cite{DBLP:journals/jmlr/GarciaF15}, as rewards often do not fully capture complex safety requirements~\cite{DBLP:journals/aamas/VamplewSKRRRHHM22}.
For example, an RL Taxi policy trained to maximize its reward for each passenger transported to their destination might not account for possible collisions.

To resolve the issue mentioned above, formal verification methods like \emph{model checking}~\cite{DBLP:conf/setta/Gross22} have been proposed to reason about the safety of RL~\cite{yuwangPCTL,DBLP:conf/formats/HasanbeigKA20,DBLP:conf/atva/BrazdilCCFKKPU14,DBLP:conf/tacas/HahnPSSTW19}.
Model checking is not limited by properties that can be expressed by rewards but supports a broader range of properties that can be expressed by \emph{probabilistic computation tree logic (PCTL)} \cite{DBLP:journals/fac/HanssonJ94}.
At its core, model checking uses mathematical models to verify a system's correctness concerning a given safety property.

Despite progress in applying verification to RL, the complexity of NNs still hides crucial details affecting safe decision-making~\cite{DBLP:journals/ml/Bekkemoen24}.
This highlights the need for research on \emph{interpretable and safety-focused RL methods} to enhance safe decision-making and promote responsible and interpretable RL development.

\emph{Pruning methods} trim NN connections to analyze their impact on performance~\cite{DBLP:conf/icdsp2/NiL22}.
Yet, they lack a focus on safety.

Therefore, by integrating model checking and NN pruning, we propose a novel method named \emph{VERINTER (VERify and INTERpret)} to \emph{exactly interpret} neuron interconnections within NN policies concerning safety measurements.
Additionally, VERINTER can be used as a safety-conscious pruning technique to eliminate unimportant connections from the NN while maintaining safety.

VERINTER takes three inputs: a \emph{Markov Decision Process (MDP)} representing the RL environment, a trained policy, and a PCTL formula for safety measurements.
We \emph{incrementally build} only the reachable parts of the MDP, guided by the trained policy~\cite{DBLP:conf/setta/Gross22}.
We then verify the policy's safety using the Storm model checker~\cite{DBLP:journals/sttt/HenselJKQV22} and the PCTL formula.

In the case of a \emph{safety violation} of the pruned RL policy, for instance, a collision likelihood above 1\%, we can extract the information that the pruned interconnections are essential for safe decision-making.
Otherwise, we may be able to prune more interconnections, only leaving the essential interconnections for safety.


Pruning an input neuron's connections removes its feature from decision-making, revealing its impact.
For instance, if pruning the passenger sensor leads to significant changes like running out of fuel in a taxi RL scenario, it highlights its critical role in safe decision-making for the trained RL policy.

Our \textbf{main contribution}, VERINTER, safely prunes NN policies with formal verification, measures the impact of specific features and NN connections on safety, and is applicable across multiple benchmarks.
\emph{This study tries to bridge the gap between formal verification and interpretable RL, creating a unified method for safe and interpretable RL policies.}

\paragraph{Related Work}
Formal verification methods for RL policies are developed~\cite{yuwangPCTL,DBLP:conf/formats/HasanbeigKA20,DBLP:conf/atva/BrazdilCCFKKPU14,DBLP:conf/tacas/HahnPSSTW19,jansen2020safe,icaart24} and 
RL policy pruning exists~\cite{DBLP:conf/iberamia/Garcia-RamirezM22,DBLP:conf/isorc/XuL0Z022,DBLP:conf/pricai/GangopadhyayDD22}.
VERINTER differs by combining both formal verification and pruning in one \emph{interpretable RL method} in the context of RL safety, accessing the impact of NN input features and connections on safety.
Gangopadhyay et al. prune NN policies focusing on reachability while we \emph{exactly verify complex safety PCTL properties} and set them into the context of \emph{interpretable RL}.
COOL-MC is a tool that verifies whether a policy violates a safety requirement or not~\cite{DBLP:conf/setta/Gross22}.
We enhance COOL-MC by integrating it with VERINTER.

\section{Background}

\paragraph{Probabilistic model checking.} A \textit{probability distribution} over a set $X$ is a function $\mu \colon X \rightarrow [0,1]$ with $\sum_{x \in X} \mu(x) = 1$. The set of all distributions on $X$ is denoted $Distr(X)$.

\begin{definition}[MDP]\label{def:mdp}
A \emph{MDP} is a tuple $M = (S,s_0,Act,Tr, rew,AP,L)$ where
$S$ is a finite, nonempty set of states; $s_0 \in S$ is an initial state; $Act$ is a finite set of actions; $Tr\colon S \times Act \rightarrow Distr(S)$ is a partial probability transition function;
$rew \colon S \times Act \rightarrow \mathbb{R}$ is a reward~function;
$AP$ is a set of atomic propositions;
$L \colon  S \rightarrow 2^{AP}$ is a labeling function.
\end{definition}
We employ a factored state representation where each state $s$ is a vector of features $(f_1, f_2, ...,f_d)$ where each feature $f_i\in \mathbb{Z}$ for $1 \leq i \leq d$ ($d$ is the dimension of the state).
The available actions in $s \in S$ are $Act(s) = \{a \in Act \mid Tr(s,a) \neq \bot\}$ where $Tr(s, a) \neq \bot$ is defined as action $a$ at state $s$ does not have a transition (action $a$ is not available in state $s$).
An MDP with only one action per state ($\forall s \in S : |Act(s)| = 1$) is a discrete-time Markov chain (DTMC) $D$.
\begin{definition}[Policy]
A memoryless deterministic policy for an MDP $M$ is a function $\pi \colon S \rightarrow Act$ that maps a state $s \in S$ to action $a \in Act$.
\end{definition}
Applying a policy $\pi$ to an MDP $M$ yields an \emph{induced DTMC} $D$ where all non-determinism is resolved.
Storm~\cite{DBLP:journals/sttt/HenselJKQV22} allows the verification of PCTL properties of induced DTMCs to make, for instance, safety measurements.

\paragraph{RL.} The standard learning goal for RL is to learn a policy $\pi$ in an MDP such that $\pi$ maximizes the accumulated discounted reward, that is, $\mathbb{E}[\sum^{N}_{t=0}\gamma^t R_t]$, where $\gamma$ with $0 \leq \gamma \leq 1$ is the discount factor, $R_t$ is the reward at time $t$, and $N$ is the total number of steps.
In RL, an agent learns through interaction with its environment to maximize a reward signal~\cite{DBLP:journals/ml/Bekkemoen24}.

\paragraph{NN policy pruning.} A NN with $d$ inputs and $|Act|$ outputs encodes a function $f \colon \mathbb{R}^d \to \mathbb{R}^{|Act|}$. 
Formally, the function $f$ is given in the form of
a sequence $\vec{W}^{(1)},\dots,\vec{W}^{(k)}$ of \emph{weight matrices} with $\vec{W}^{(i)} \in \mathbb{R}^{d_i \times d_{i-1}}$, for all $i = 1,\dots,k$.
Pruning a weight $\vec{W}^{(k)}_{ij}$ sets it to zero, eliminating the connection between neuron $i$ in layer $k$ and neuron $j$ in layer $k+1$.

We focus on the following types of pruning: 
\emph{$l_1$-pruning} removes a specific fraction \( p \) of the weights \( \vec{W}^{(k)}_{ij} \) starting with those of the smallest $l_1$-magnitude in layer \( k \); 
\emph{Random pruning} randomly eliminates a fixed fraction $p$ of weights $\vec{W}^{(k)}_{ij}$; 
\emph{Feature pruning} cuts all outgoing connections $\vec{W}^{(1)}_{ij}$ from a neuron linked to a specific NN policy observation feature~$f_i$.

\section{Methodology}
We introduce \emph{VERINTER's workflow}, where we first incrementally build the induced DTMC of the policy $\pi$ and the MDP $M$ as follows. For every reachable state $s$ via the trained policy $\pi$, we query for an action $a = \pi(s)$. In the underlying MDP $M$, only states $s'$ reachable via that action $a \in A(s)$ are expanded. The resulting DTMC $D$ induced by $M$ and $\pi$ is fully deterministic, with no open action choices, and is passed to the model checker Storm for verification, yielding the \emph{exact} safety measurement result $m$.

Next, the pruning procedure eliminates connections $\hat{W}$ within the NN based on predefined criteria and verifies the induced DTMC $\hat{D}$ of the pruned policy $\hat{\pi}$ and the MDP $M$ to obtain the measurement result $\hat{m}$. Our framework remains independent of the specific pruning method used.

Then, with the completion of the pruning process, we can examine the difference between $m$ and $\hat{m}$ to evaluate the relevance of the pruned connections.

\paragraph{Safety feature pruning}
A feature $f_i$ is important for an RL policy $\pi$ and a specific measurement if its removal impacts policy safety.
For instance, if all outgoing connections from the input layer receiving feature $f_i$ are pruned, resulting in a pruned RL policy $\hat{\pi}$, we can assess if $f_i$ is crucial for safety performance.
In a taxi scenario where a passenger sensor is removed, and safety performance remains unaffected, such as the likelihood of running out of fuel remaining unchanged, we deduce that this feature does not influence this safety.

\paragraph{Limitations}
Our method supports memoryless NN policies within modeled MDP environments, only limited by state space and transition count~\cite{DBLP:conf/setta/Gross22}. VERINTER remains independent of the pruning method.

\section{Experiments}
We evaluate VERINTER in multiple model-based environments from~\cite{DBLP:conf/setta/Gross22} (\emph{Taxi, Freeway, Crazy Climber, Avoidance, and Stock Market}).
Experiments involve training RL policies using the deep Q-learning algorithm~\cite{mnih2013playing}, achieving high safety success across the environments. The \emph{Taxi policy} maintains non-empty full status, completes two jobs, and reaches a gas station with 100\% success; the \emph{Freeway policy} crosses 100\% of the time safely; the \emph{Crazy Climber policy} avoids falls; the \emph{Avoidance policy} prevents collisions 68\% of the time; and the \emph{Stock Market policy} avoids bankruptcy.

\paragraph{Comparative analysis of pruning methods.}
\begin{figure}[] 
    \centering
    \includegraphics[width=0.6\columnwidth]{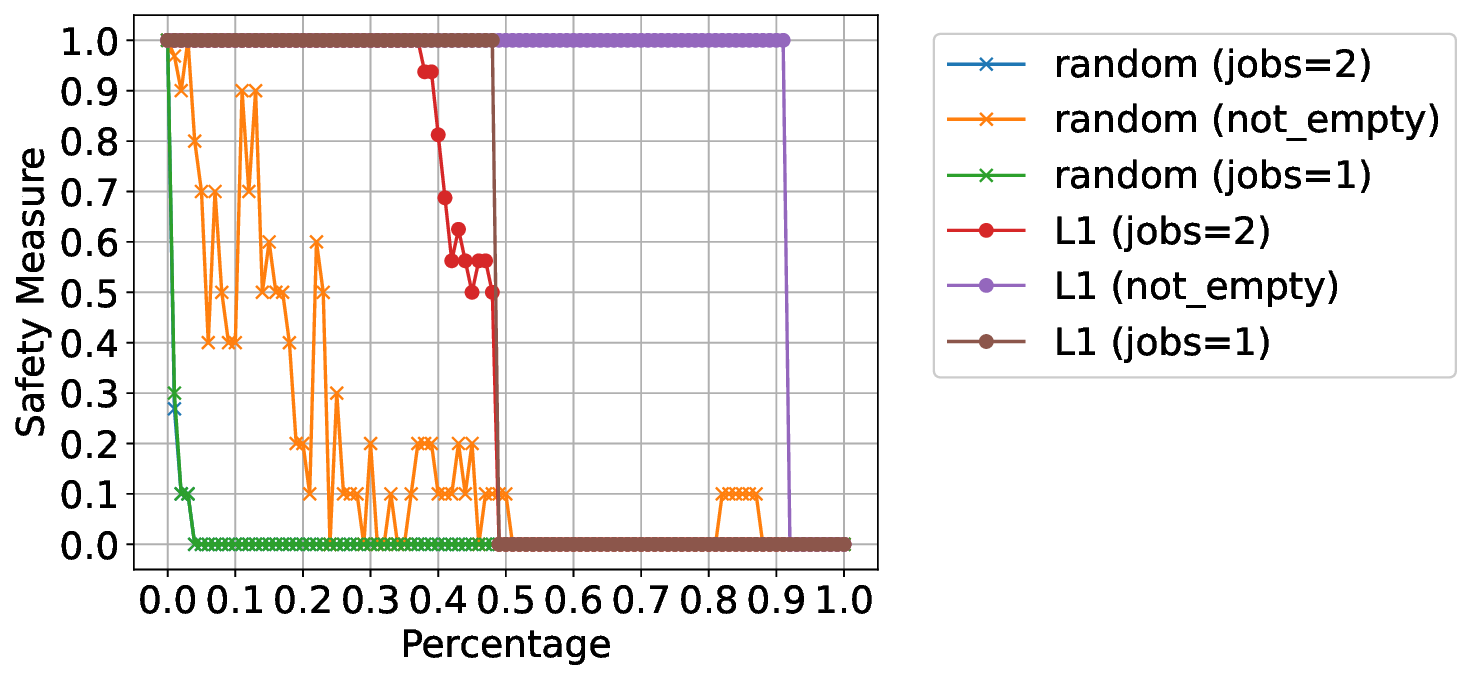}
    \caption{Pruning methods across three Taxi events. The x-axis shows the percentage of pruned weights in the input layer, and the y-axis indicates the reachability probability of specified events (in brackets). Random pruning sample size: 10.} 
    \label{fig:pruning_compare} 
\end{figure}
This experiment compares two pruning methods on $W^{(1)}$, highlighting the model-agnostic nature of our method.
In Figure~\ref{fig:pruning_compare}, $l_1$-pruning removes more connections than random pruning while maintaining initial safety performance.
Random pruning lacks consideration of connection weight, risking the removal of crucial connections and causing rapid performance degradation.
Therefore, different pruning methods uniquely affect safety measurements due to varying connection pruning strategies.
\begin{figure}[]
  \centering
  \begin{subfigure}{0.5\columnwidth} 
    \centering
    \includegraphics[width=1\textwidth]{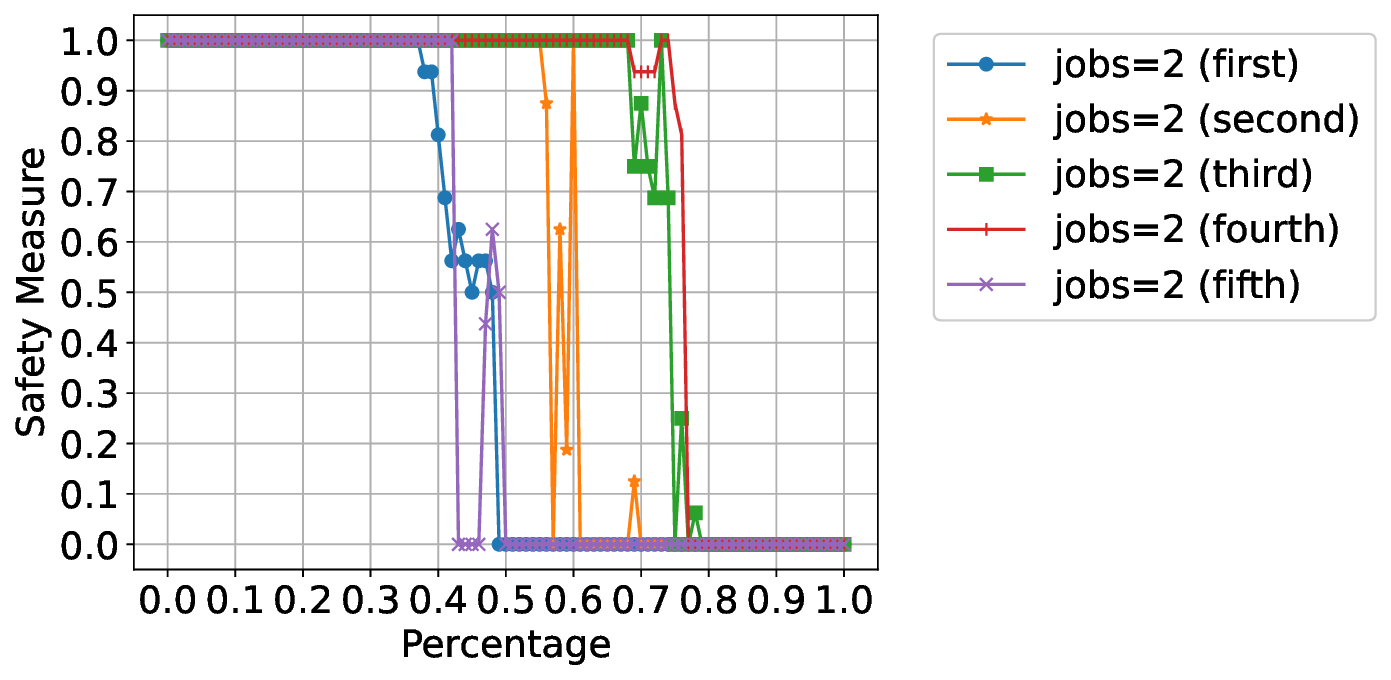}
    \caption{The reachability probability for finishing two jobs.} 
    \label{fig:subfig1} 
  \end{subfigure}
  \begin{subfigure}{0.45\columnwidth} 
  \centering
    \includegraphics[width=1.1\textwidth]{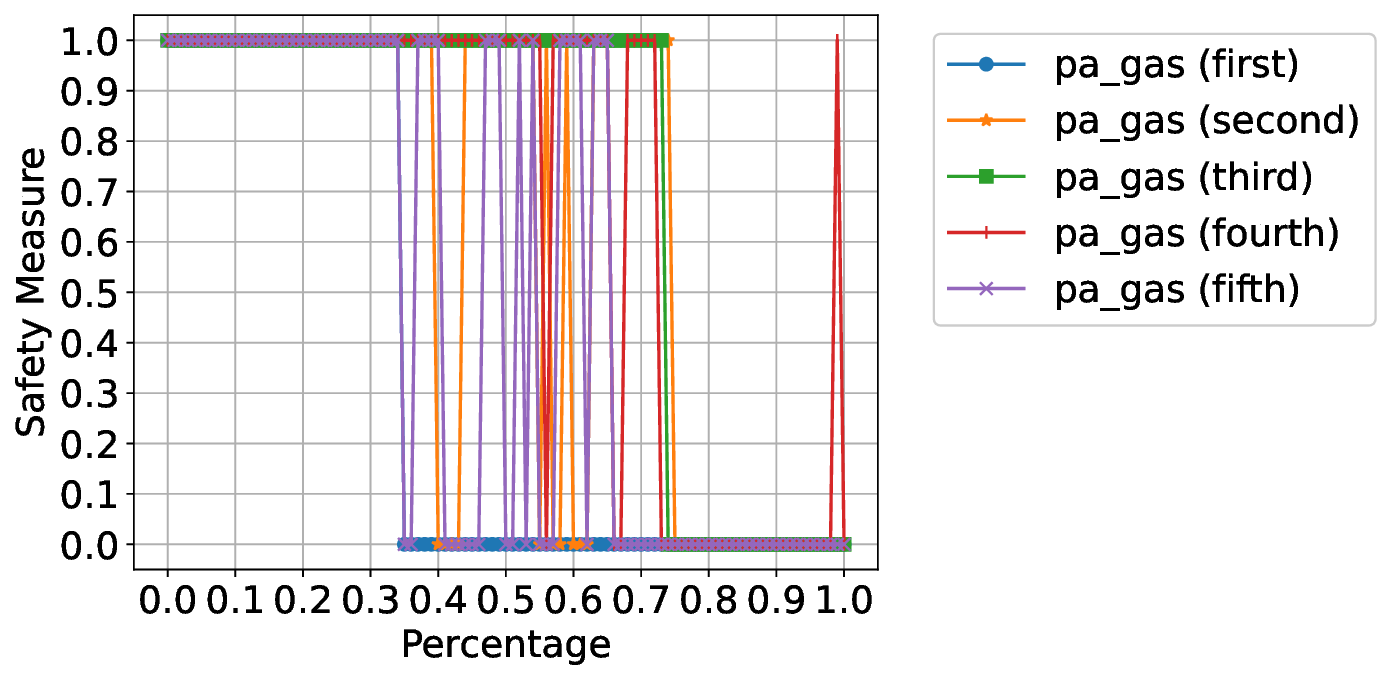}
    \caption{The probability of picking up a passenger and then visiting a gas station.} 
    \label{fig:subfig3} 
  \end{subfigure}
  \caption{Each subfigure shows safety measurements for different NN layers, with the x-axis representing the percentage of pruned connections and the y-axis showing safety outcomes in the Taxi environment.} 
  \label{fig:different_layers} 
\end{figure}

\paragraph{Effect of pruning different layers.}
We examine how pruning different layers of a trained NN policy affects safety in the Taxi scenario. We focus on safety measurements for completing two jobs (\emph{jobs=2}) and the \emph{complex probability measurement}~\cite{DBLP:journals/aamas/VamplewSKRRRHHM22} of picking up the passenger before reaching the gas station (\emph{pa\_gas}).
In Figure~\ref{fig:different_layers}, the pruning impact on safety does not consistently relate to specific layers, suggesting no dominance of low-level layers over high-level ones.
Notable, pruning the first layer (shown by the blue line) slightly increased the reachability probability of completing two jobs with around 42\% of neurons pruned, indicating that (further) pruning could enhance safety performance.
\begin{table}[t]
\centering
\small
\begin{tabular}{@{}llrlr@{}}
\toprule
\textbf{Environment}  & \textbf{Measure Label} & \textbf{Orig. Result} & \textbf{Pruned Feature} & \textbf{Result} \\
\midrule
Taxi  & $jobs=2$ & $1$ & passenger\_loc\_x &  $1$ \\
Freeway  & $crossed$ & $0.99$ & $px_0$ &  $0.98$ \\
Crazy Climber  & $no\_fall$ & $0$ & $px_1$ &  $0$ \\
Avoidance  & $no\_collision_{100}$ & $0.68$ & $x$ &  $0.25$ \\
Stock Market  & $no\_bankruptcy$ & $1$ & $sell\_price$ &  $1$ \\
\bottomrule
\end{tabular}
\caption{Safety feature prunings. The \textit{Measure Label} refers to the safety measure. The \textit{Orig. Result} and \textit{Result} show the probability of conformance to the safety measure before and after pruning of the input feature \textit{Pruned Feature}.}
\label{table:different_benchmarks}
\end{table}

\paragraph{Safety feature pruning in different environments.}
Our method adapts to different RL environments, as shown in Table~\ref{table:different_benchmarks}. The results vary; some pruned features maintain safety, while others compromise it.

\section{Conclusion}
VERINTER integrates model checking with NN pruning to refine RL policies, maintaining performance while identifying expendable features and NN connections. \emph{Future research} could include multi-agent RL~\cite{DBLP:journals/aamas/ZhuDW24}.

\begin{footnotesize}
\bibliographystyle{unsrt}
\bibliography{sample-base}

\end{footnotesize}

\end{document}